\documentclass{article}

\usepackage{arxiv}
\usepackage[utf8]{inputenc} % allow utf-8 input
\usepackage[T1]{fontenc}    % use 8-bit T1 fonts
\usepackage{hyperref}       % hyperlinks
\usepackage{url}            % simple URL typesetting
\usepackage{booktabs}       % professional-quality tables
\usepackage{amsfonts}       % blackboard math symbols
\usepackage{nicefrac}       % compact symbols for 1/2, etc.
\usepackage{microtype}      % microtypography
\usepackage{lipsum}		% Can be removed after putting your text content
\usepackage{graphicx}
\usepackage{natbib}
\usepackage{doi}

\usepackage{mathtools}
\usepackage{amssymb}
\usepackage{graphicx}
\usepackage{epstopdf}
\usepackage{times}
\usepackage{epsfig}
\usepackage{subfigure}
\usepackage[utf8]{inputenc}
\usepackage{amsthm}
\usepackage{algorithm}
\usepackage{algorithmic}
\usepackage{xcolor}
\usepackage{booktabs}
\usepackage{multirow}

\newtheorem{definition}{Definition}

\title{Multi-channel Deep 3D Face Recognition}

\date{January 15, 2020}	
\author{ 
		{\hspace{1mm}Zhiqian You} \\
	Center for Advanced Computer Studies\\
	School of Computing and Informatics\\
	University of Louisiana at Lafayette\\
	Lafayette, LA 70503 \\
	\texttt{zhiqianyzq@gmail.com } \\
	\And
		{\hspace{1mm}Tingting Yang} \\
	Center for Advanced Computer Studies\\
	School of Computing and Informatics\\
	University of Louisiana at Lafayette\\
	Lafayette, LA 70503 \\
	\texttt{tingting.yang1@louisiana.edu} \\
	\And
	{\hspace{1mm}Miao Jin} \\
	Center for Advanced Computer Studies\\
	School of Computing and Informatics\\
	University of Louisiana at Lafayette\\
	Lafayette, LA 70503 \\
	\texttt{miao.jin@louisiana.edu} \\
}

\renewcommand{\shorttitle}{\textit{arXiv} Template}

\begin{document}
\maketitle

\begin{abstract}
	Face recognition has been of great importance in many applications as a biometric for its throughput, convenience, and non-invasiveness. Recent advancements in deep Convolutional Neural Network (CNN) architectures have boosted significantly the performance of face recognition based on two-dimensional (2D) facial texture images and outperformed the previous state of the art using conventional methods. However, the accuracy of 2D face recognition is still challenged by the change of pose, illumination, make-up, and expression. On the other hand, the geometric information contained in three-dimensional (3D) face data has the potential to overcome the fundamental limitations of 2D face data. 
	
	We propose a multi-Channel deep 3D face network for face recognition based on 3D face data. We compute the geometric information of a 3D face based on its piecewise-linear triangular mesh structure and then conformally `flatten' geometric information along with the color from 3D  to 2D plane to leverage the state-of-the-art deep CNN architectures. We modify the input layer of the network to take images with nine channels instead of three only such that more geometric information can be explicitly fed to it.  We pre-train the network using images from the VGG-Face~\cite{Parkhi2015} and then fine-tune it with the generated multi-channel face images. The face recognition accuracy of the multi-Channel deep 3D face network has achieved $98.6\%$. The experimental results also clearly show that the network performs much better when a 9-channel image is `flattened' to plane based on the conformal map compared with the orthographic projection.
\end{abstract}

% keywords can be removed

\section{Introduction}

Face recognition has been of great importance in many applications as a biometric for its throughput, convenience, and non-invasiveness. It has been an active research topic for years with the challenging of varying facial appearance due to changes in pose, illumination, make-up, expression, or hard occlusions.

Recent advancements in deep Convolutional Neural Network (CNN) architectures have boosted significantly the performance of face recognition based on two-dimensional (2D) facial texture images~\cite{DeepFace2014, Parkhi2015, FaceNet2015}. Deep networks integrate different levels of facial features and recognize them in an end-to-end fashion. They outperform the previous state of the art using conventional methods with hand-crafted feature extractors including Local Binary Pattern~\cite{Ahonen_facerecognition2004} and Fisher vectors~\cite{Simonyan13}.

However, the accuracy of face recognition based on 2D facial texture image is still challenged by the change of pose, illumination, make-up, and expression. It is highly desirable to input a CNN network with more robust information. Therefore, the geometric information contained in a three-dimensional (3D) face point cloud has the potential to overcome the fundamental limitations of 2D facial texture images. 

Pioneering works have been done in~\cite{DBLP:journals/corr/KimHCM17,DBLP:journals/corr/abs-1711-05942}, which apply deep CNN architectures for 3D face recognition. In both works, a 3D face point cloud is projected onto a 2D image plane with orthographic projection, called the depth image. Kim et al.~\cite{DBLP:journals/corr/KimHCM17} feed the VGG-Face network~\cite{Parkhi2015} with an augmented 3D face dataset consisting of 123,325 depth images. They test the network on three datasets: Bosphorus~\cite{Bosphorus}, BU3DFE~\cite{BU3DFE}, and 3D-TEC (twins)~\cite{3D-TEC}. Their results outperform the state-of-the-art conventional 3D face recognition methods in the Bosphorus dataset.

Considering a depth image occupies only one channel, Gilani et al.~\cite{DBLP:journals/corr/abs-1711-05942} reserve another two channels for surface normals represented by spherical coordinates $(\theta, \phi)$ to provide more geometric information. Normals are calculated on the original 3D point cloud and then projected to a 2D image plane with orthographic projection. The three channels are then normalized on the $0-255$ range and rendered as an RGB image to feed to the Deep 3D Face Recognition Network (FR3DNet) with a skeleton architecture similar to~\cite{Parkhi2015}. The authors generate millions of 3D facial images for training by simultaneously interpolating between the facial identities and expression spaces. Results of FR3DNet outperform the accuracy of all conventional 3D face recognition methods of existing 3D face datasets. 

The primary downside of the two existing methods is that there is far richer geometric information on a surface that represents the features of a face and the geometric information orthographically projected from 3D to 2D has been largely distorted.  

The multi-Channel deep 3D face network is completely different from the above works in that: (1) Geometric information of a 3D face is conformally mapped from 3D to 2D plane. A conformal map preserves surface angles and local shapes everywhere. (2) The network accepts images with multi-channel (more than three channels) that contain far richer geometric information. Specifically, we convert the 3D face point cloud to a  piecewise-linear triangular mesh and then compute the geometric information of the 3D face based on the triangular mesh. The computed geometric information along with the face color is conformally mapped to a 2D plane as a multi-channel face image to leverage state-of-the-art deep CNN architectures. Note that the conformal map is intrinsic to the geometry of a 3D face, independent of its triangulation resolution.  We then modify the input layer of deep CNN architectures to take images with more channels instead of just three. The multi-Channel deep 3D face network is trained first using images from the VGG-Face~\cite{Parkhi2015} and then fine-tuned with the multi-channel face images.  

The rest of the paper is organized as follows: Section~\ref{sec:image} explains the way to generate a multi-channel face image from 3D face point cloud data. Section~\ref{sec:Multi-Channel} introduces the multi-channel deep 3D face network. Section~\ref{sec:experiments} gives the experimental results. We conclude the paper in Section~\ref{sec:conclusion}.

\section{Multi-Channel Face Image}\label{sec:image}

A scanned 3D face is stored as either a depth file or point cloud. It is straightforward to convert a depth file to a triangular mesh. For a point cloud, we re-sample the points and then convert them to a triangular mesh. 

The rich geometric information of a 3D face can be computed based on its piecewise-linear triangular mesh structure. However, we need a tool to convert the computed geometric information and the stored color at each vertex from 3D to 2D. A simple orthographic projection in~\cite{DBLP:journals/corr/KimHCM17,DBLP:journals/corr/abs-1711-05942} drops $z$ dimensional information in a brute force way. The implementation is simple, but the cost is high, which puts 3D face recognition vulnerable under different poses as its 2D counterpart. 

A conformal map preserves surface angles and local shape. The mapping itself is intrinsic to the geometry of a 3D face, independent of its triangulation. These properties make a conformal map an ideal choice to convert the computed geometric information from 3D to 2D, convenient for CNN architectures. 

We explain briefly the concept of the conformal map in Sec.~\ref{sec:conformal}, the tool we use to compute the conformal map in Sec.~\ref{sec:Ricci}, and the geometric properties we compute in Sec.~\ref{sec:geometry}. 

\subsection{Conformal Map}\label{sec:conformal}

A \emph{conformal map\index{conformal map}} between two surfaces preserves angles. Specifically, let $S$ be a surface embedded in $\mathbb{R}^3$ with a Riemannian metric induced from the Euclidean metric of $\mathbb{R}^3$, denoted by $\mathbf{g}$. Suppose $u:S\rightarrow\mathbb{R}$ is a scalar function defined on $S$. It can be verified that $\bar{\mathbf{g}}=e^{2u}\mathbf{g}$ is also a Riemannian metric on $S$. Furthermore, angles measured by $\mathbf{g}$ are equal to those measured by $\bar{\mathbf{g}}$. Therefore, we say $\bar{\mathbf{g}}$ is a \emph{conformal	deformation} from $\mathbf{g}$. \emph{Riemann mapping} theorem states that any simply connected surface with a single boundary, i.e., a topological disk, can be conformally mapped to a unit disk. 

\begin{figure*}[h!]
	\centering
	\begin{tabular}{ccc}	
		\includegraphics[height=4.0cm]{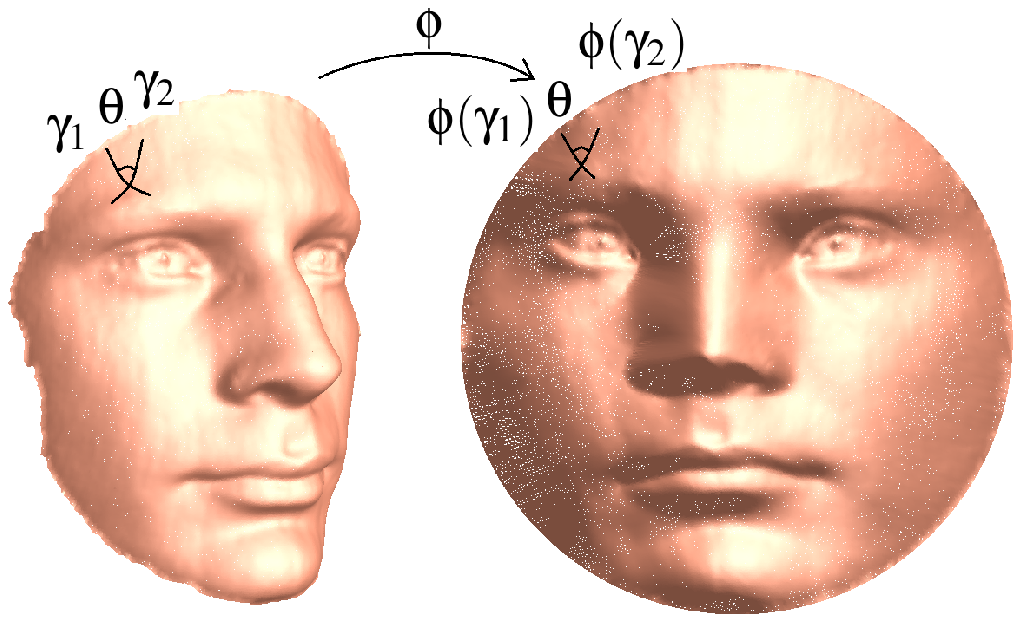} &
		\includegraphics[height=4.0cm]{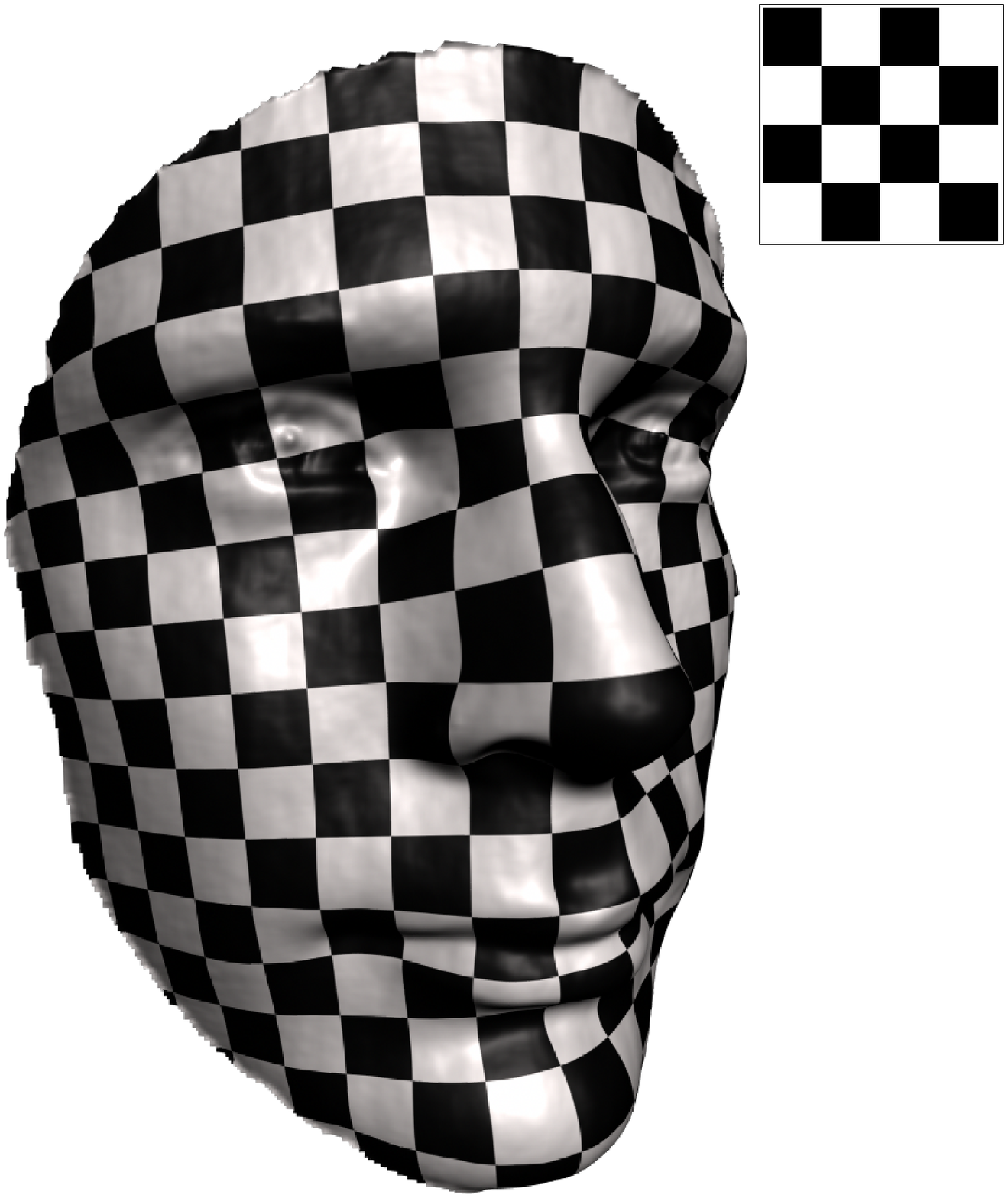} &
		\includegraphics[height=4.0cm]{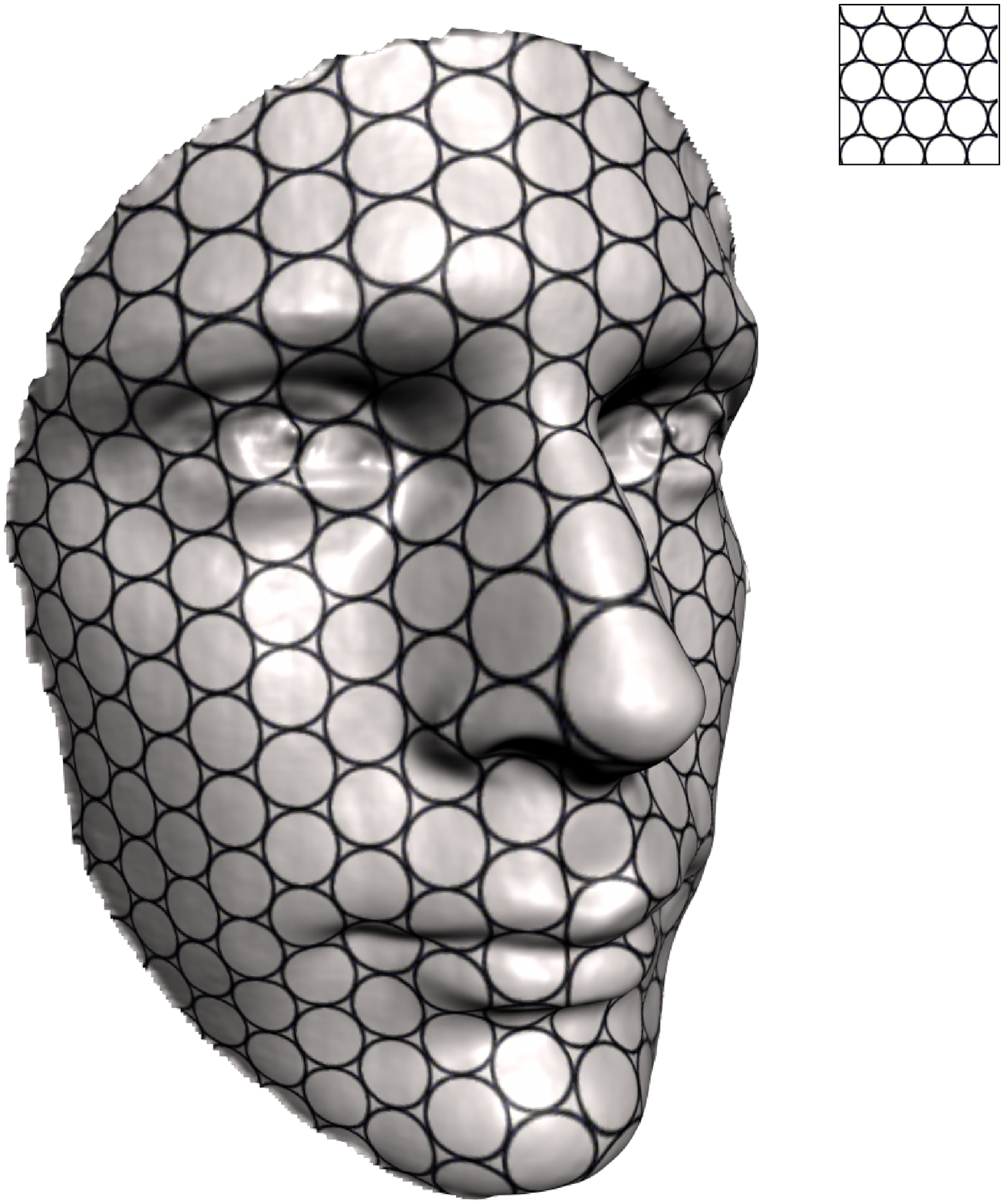} \\
		(a) & (b) & (c) \\
	\end{tabular}
	\caption{Visualization of a conformal map with texture mapping:	(a) A 3D face surface is conformally mapped to a unit disk. (b) A planar checkerboard texture is pulled back to the 3D surface based on the inverse of the conformal mapping where all the squares including their right angles are well preserved on the surface. (c) A circle packing texture is pulled back to the 3D surface where all the circles including their tangency relations are well preserved on the surface. } \label{fig:vis_conformality}
\end{figure*}

Figure~\ref{fig:vis_conformality} (a) shows a scanned 3D human face, i.e., a topological disk surface denoted as $S$, mapped to a unit disk denoted as $D$ by $\phi: S\to D$. Suppose $\gamma_1,\gamma_2$ are two arbitrary curves on the face surface $S$, and $\phi$ maps them to $\phi(\gamma_1),\phi(\gamma_2)$. If the intersection angle between $\gamma_1,\gamma_2$ is $\theta$, then the intersection angle between $\phi(\gamma_1)$ and $\phi(\gamma_2)$ is also $\theta$. $\gamma_1$ and $\gamma_2$ can be chosen arbitrarily. Therefore, we say $\phi$
is conformal, meaning angle-preserving.

A conformal deformation maps infinitesimal circles to infinitesimal circles and preserves the intersection angles among them, so locally a conformal map introduces no distortion, only scaling.  Figures~\ref{fig:vis_conformality} (b) and (c) visualize the properties based on texture mapping technique. Texture refers to an image on the plane. Based on the conformal map shown in Figure~\ref{fig:vis_conformality} (a), we cover the planar disk by a checkerboard texture image and then pull back the image onto the 3D face surface. Since the mapping is conformal, all the squares including their right angles of corners are well preserved on the human face as shown in Figure~\ref{fig:vis_conformality} (b). If we replace the
texture with a circle packing pattern, then planar circles are mapped to circles on the surface. All the circles including their tangency relations are well preserved as shown in Figure~\ref{fig:vis_conformality} (c).

\subsection{Discrete Surface Ricci Flow}\label{sec:Ricci}

Richard Hamilton first introduced Ricci flow in his seminal work~\cite{ric82}. Chow and Luo in~\cite{chow_luo_03} proved a general existence and convergence theorem for the discrete Ricci flow on surfaces. Later computational algorithms of discrete surface Ricci flow are provided in~\cite{TVCG_08_Ricci}. 

Discrete surface Ricci flow is a powerful tool to compute surface conformal deformation with the flexibility to design target Gaussian curvatures. To briefly introduce the concept of discrete surface Ricci flow, we start from the definition of discrete metric, discrete Gaussian curvature, and circle packing metric.

In discrete setting, we denote $M= (V, E, F)$ a connected triangular mesh embedded in $\mathbb{R}^3$, consisting of vertices ($V$), edges ($E$), and triangle faces ($F$). Specifically, we denote $v_i \in V$ a vertex, $e_{ij} \in E$ an edge with two ending vertices $v_i$ and $v_j$; $f_{ijk} \in F$ a triangle face with vertices $v_i$, $v_j$, and $v_k$.  

\begin{definition} [Discrete Metric] A discrete metric on $M$ is a function $l: E \to \mathbb{R}^+$ on the set of edges, assigning to each edge $e_{ij}\in E$ a positive number $l_{ij}$ such that the triangle inequalities are satisfied for all triangles $t_{ijk} \in F$: $l_{ij} + l_{jk} > l_{ki}$.
\end{definition}

The edge lengths of a triangular mesh $M$ are sufficient to define its discrete metric.

\begin{definition}[Discrete Gaussian Curvature] The discrete Gaussian curvature $K_i$ on a vertex $v_i\in V$ can be computed from the angle deficit:
	\begin{equation}
	K_{i}= \left\{
	\begin{array}{rl}
	2\pi - \sum_{f_{ijk} \in F} \theta_{i}^{ij},
	& v_i \not\in \partial M, \\
	\pi - \sum_{f_{ijk} \in F} \theta_{i}^{jk},
	& v_i \in \partial M,
	\end{array}
	\right.
	\label{eqn:discrete_gaussian_curvature}
	\end{equation}
	where $\theta _{i}^{jk}$ represents the corner angle attached to Vertex $v_i$ in Face $f_{ijk}$ and $\partial M$ is the boundary of the mesh.
\end{definition}

It is obvious that the discrete Gaussian curvatures can be fully computed from the discrete metric.

Assign each vertex $v_i$ a circle with radius $\gamma_i$. We  denote the radius function as $\Gamma : V \to \mathbb{R}^{+}$. The two circles centered at vertices $v_i$ and $v_j$, respectively of edge $e_{ij}$ intersect with an acute angle $\phi_{ij}$. We call angle $\phi_{ij}$ the \emph{weight} on edge $e_{ij}$. We  denote the edge weight function as $\Phi: E \to [0,\frac{\pi}{2}]\label{eqn:edge_weight}$.

The length of an edge $e_{ij}$ can be computed from the vertex circle radii $\gamma_i,\gamma_j$ and the weight $\phi_{ij}$ by the following cosine law:
\begin{equation}
{l_{e_{ij}}}^2 = {\gamma_i}^2 + {\gamma_j}^2 + 2\gamma_i \gamma_j\cos\phi_{ij}.
\label{eqn:cosine}
\end{equation}

Thurston introduced the circle packing metric~\cite{Thurston76}:
\begin{definition}[Circle Packing Metric]
	A circle packing metric of a mesh $M$ includes the circle radius function and the edge weight function.
\end{definition}

\begin{definition}[Discrete Conformal Deformation] Two circle packing metrics $(\Gamma_1,\Phi_1)$ and $(\Gamma_2,\Phi_2)$ on the same mesh are \emph{conformally equivalent} if $\Phi_1 \equiv \Phi_2$. A \emph{conformal	deformation} of a circle packing metric  modifies the vertex radii but preserves their intersection angles.
\end{definition}

\begin{definition}[Discrete Surface Ricci Flow] Suppose mesh $M$ has an initial circle packing metric $(\Gamma_0,\Phi)$. Let $u_i$ be the logarithm of $\gamma_i$ associated with vertex $v_i$. Discrete surface Ricci flow is defined as follows:
	\begin{equation}
	\frac{du_i(t)}{dt} = (\bar{K}_i-K_i ),\\
	\label{eqn:discrete_ricci_flow}
	\end{equation}
	where $\bar{K}_i$ and $K_i$ are the target and current Gaussian curvatures at $v_i$ and $t$ is the time. Discrete surface Ricci flow  deforms the circle packing metric according to the difference of the current and target Gaussian curvatures. The final circle packing metric induces the metric that satisfies the target Gaussian curvature.
\end{definition}

Discrete surface Ricci flow is a negative gradient flow of a special energy form, the so called \emph{discrete
	Ricci energy}:
\begin{equation}
f( \mathbf{u} )=\int_{(\Gamma_0,\Phi)}^{(\Gamma,\Phi)} \sum_{i=1}^n(\bar{K_i}-K_i) du_i, \label{eqn:discrete_ricci_energy}
\end{equation}
where $(\Gamma_0,\Phi)$ is the initial circle packing metric, which induces the surface original metric.
It has been shown in \cite{chow_luo_03} that the discrete Ricci energy is convex with a unique global minimum. The minimum corresponds to the desired metric $(\Gamma,\Phi)$, which induces the target Gaussian curvature. Discrete surface Ricci flow converges to this unique global minimum with an exponentially fast convergence speed that can be estimated by the following formula \cite{chow_luo_03}:
\[
|K_i(t) - \bar{K}_i | < c_1e^{-c_2 t}, c_1, c_2 > 0.
\]

Since the boundary shape of a mapped 3D face mesh on a plane won't affect the face recognition, and a fixed boundary shape, e.g., a unit disk shape,  will bring large area distortion to boundary regions, we apply the computational discrete surface Ricci flow algorithm~\cite{TVCG_08_Ricci} with free-boundary shape condition. Specifically, we assign the target Gaussian curvatures of all interior vertices to zero and discrete surface Ricci flow deforms only non-boundary edge lengths. The algorithm is as follows:

\begin{enumerate}
	\item  Initialization of circle packing metric: compute the initial circle packing metric based on the edge lengths of the input 3D face mesh denoted as $M$.
	
	\item  Initialization of target Gaussian curvature: set the target Gaussian curvature of all non-boundary vertices  $ \bar{K_i} = 0 $ where $v_i \not\in \partial M$.
	
	\item  Set $\epsilon$, the threshold of the curvature error between the current and target Gaussian curvatures. 
	
	\item Apply the algorithm in~\cite{TVCG_08_Ricci} to compute the desired flat metric. Specifically, discrete surface Ricci flow deforms the circle packing metric of non-boundary vertices until all $|\overline{k_i} - k_i| < \epsilon$ where $v_i \not\in \partial M$.
	
	\item  Planar embedding: When the discrete surface Ricci flow converges, the final circle packing metric determines the edge lengths (i.e., $\{l_{ij}|e_{ij}\in E\}$) of $M$ mapped in plane. Starting from one $f_{ijk}$, set their $uv$ values as: $uv(v_i)=(0,0)$, $uv(v_j)=(l_{ij},0)$, and $uv(v_k)=(l_{ki}cos\theta_{i}^{jk}, l_{ki}cos\theta_{i}^{jk})$. In a breadth first search way, for $f_{jil}$ with exactly two vertices (e.g., $v_i$ and $v_j$) having $uv$ values, compute the $uv$ value of $v_l$ as the intersection point of the two circles centered at $uv(v_i)$ and $uv(v_j)$ with radii $l_{il}$ and $l_{jl}$, respectively, and satisfying $(uv(v_l)-uv(v_i)) \times (uv(v_j)-uv(v_l)) > 0$. Repeat the above process until every vertex has the $uv$ value, i.e., its planar coordinates.
\end{enumerate}

Figure~\ref{fig:architecture} (b) shows the mapped 3D face mesh on a plane under free-boundary condition. 

\subsection{Geometric Properties}\label{sec:geometry}

Due to the piecewise-linear nature of a triangular mesh, the notions of normal, curvatures, their derivatives, and other differential properties of surfaces,  well known in Differential Geometry~\cite{Gray:2006}, become nontrivial.  We refer readers to \cite{Meyer02discretedifferential-geometry,Cohen-Steiner:2003:RDT:777792.777839,Rusinkiewicz:2004:ECD:1018408.1018660,Gatzke06estimatingcurvature} for detailed and more accurate approximation of these differential properties. 

Normal vector at vertex $v_i$ is estimated as a weighted average of the normals of the triangle faces incident to $v_i$:
\begin{equation}
Nor_i = \frac{\sum_{i} \alpha_i A_i N_i}{||\sum_{i} \alpha_i A_i N_i||},
\end{equation}
where $\alpha_i$, $A_i$, and $N_i$ represent the corner angle, area, and normal of triangle attached to Vertex $v_i$, respectively.

Gaussian curvature at vertex $v_i$ is approximated as the following weighted angle deficit:
\begin{equation}
K_{i}= \left\{
\begin{array}{rl}
\frac{3}{2\sum_{i} A_i}(2\pi - \sum_{i} \alpha_i),
& v_i \not\in \partial M, \\
\frac{3}{2\sum_{i} A_i}(\pi - \sum_{i} \alpha_i),
& v_i \in \partial M,
\end{array}
\right.
\label{eqn:discrete_gaussian_curvature}
\end{equation}
where $A_i$ and $\alpha_i$ represent the area and corner angle of triangle attached to Vertex $v_i$, respectively, and $\partial M$ is the boundary of the mesh.

Conformal mapping locally introduces no other distortion, only scaling. Such scaling is called conformal factor. Conformal factor at vertex $v_i$ can be approximated as the ratio of the sum of the areas of triangles incident to $v_i$ in $3D$ and $2D$ plane:
\[
cf_i = \frac{ \sum_{i} A_i^{3D} }{ \sum_{i} A_i^{2D} },
\]
where $A_i^{3D}$ and $A_i^{2D}$ represent the areas  of triangle attached to Vertex $v_i$ in 3D and 2D, respectively.

\section{Multi-Channel Deep 3D Face Network}\label{sec:Multi-Channel}

\subsection{Preprocessing}

We pre-train the multi-Channel deep 3D face network using images from the VGG-Face~\cite{Parkhi2015} and then fine-tune the network with the generated multi-channel face images as introduced in Sec.~\ref{sec:image}. For all images, we apply Multi-task Cascaded Convolutional Networks introduced in~\cite{DBLP:journals/corr/ZhangZL016} for face alignment. The framework adopts a cascaded
structure with three stages of deep convolutional networks that predict the face and landmark locations in a coarse-to-fine manner and achieves very good performance for alignment. All the aligned images are resized to $182 \times 182$.

\subsection{Network Architecture}

Residual Networks (ResNets) introduced in~\cite{DBLP:conf/cvpr/HeZRS16} have achieved impressive, record-breaking performance in ImageNet~\cite{Russakovsky:2015:ILS:2846547.2846559}. Training this form of networks has been shown to be easier than training plain deep convolutional neural networks. The problem of degrading accuracy can also be resolved.

We use ResNet34 as the core architecture to minimize the triplet loss defined in~\cite{FaceNet2015}:
\begin{equation}
\sum_{i}^{N} ||f(x_i^a) - f(x_i^p)||_2^2 - ||f(x_i^a) - f(x_i^n)||_2^2 + \alpha,
\label{eqn:triplet_loss}
\end{equation}

where $f(x) \in R^d$ represents the embedding of an image $x$ into a $d-dimensional$ Euclidean space,  $x_i^a$, $x_i^p$, and $x_i^n$ represent a specific person (the anchor image), another image of the same person (the positive image), and any other person (the negative image) respectively, and $\alpha$ is a margin that is enforced between positive and negative pairs. Note that  $N=9$ with 9 channels for each image in the multi-Channel deep 3D face network. We set the dimension $d$ to $128$ and the margin $\alpha$ to $0.5$.

\begin{figure*}[h!]
	\centering
	\begin{tabular}{c}
		\includegraphics[height=2.8cm]{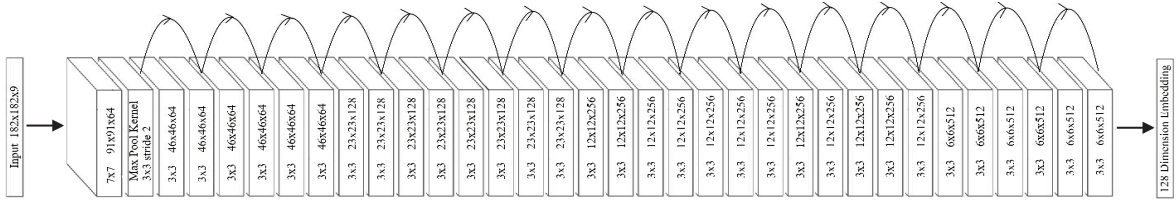} \\
		(a) Architecture of the multi-Channel deep 3D face network \\
		\\
		\includegraphics[height=1.8cm]{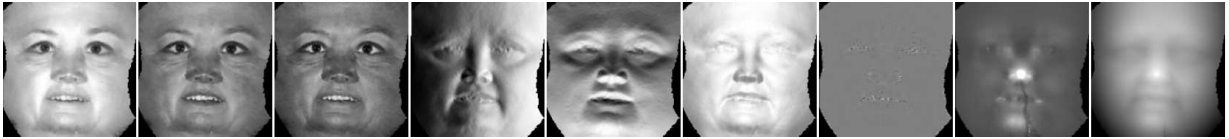} \\
		\\
		(b) An example of the 9-channel image \\
	\end{tabular}
	\caption{ (a) ResNet34 is the core architecture of the multi-Channel deep 3D face network where a shortcut connection is added every two convolutional layers. The first convolutional layer takes images with $9$ channels and the filter size is $7x7$. Note that each convolutional layer is followed by a batch	normalization (BN)~\cite{Ioffe:2015:BNA:3045118.3045167} and a ReLU layer both of that are not shown in the figure.   (b) An example of the 9-channel image: the nine channels correspond to the R, G, B colors, the three dimensions of a normal vector, the Gaussian curvature, the conformal factor, and the depth information respectively. }
	\label{fig:architecture}
\end{figure*}

Figure~\ref{fig:architecture} (a) shows the skeleton structure of the multi-Channel deep 3D face network where a shortcut connection is added every two convolutional layers. The first convolution layer takes images with $9$ channels and the kernel size is $7 \times 7$ as suggested in~\cite{DBLP:journals/corr/abs-1711-05942}. Note that each convolutional layer is followed by a batch	normalization (BN)~\cite{Ioffe:2015:BNA:3045118.3045167} and a ReLU layer both of that are not shown in Figure~\ref{fig:architecture}. 

\subsection{Implementation}

We implement the multi-Channel deep 3D face network using PyTorch. We then start from VGG-Face~\cite{Parkhi2015} with $1,648,187$ images and $2,613$ subjects in total to pre-train the network from scratch. The input to the network is $182 \times 182 \times 9$ image
where the nine channels correspond to repeated R, G, B colors, respectively. We initialize the weights as in~\cite{He:2015:DDR:2919332.2919814} and then optimize the learning using Stochastic Gradient Descent (SGD) with a mini-batch size of $128$ and two GPUs. The learning rate starts from $0.01$, reduced by a factor of $10$ after every $50$ epochs. The models are trained for up to $150$ epochs. We use a weight decay of $0.0001$ and a momentum of $0.9$.

We then fine-tune the network with the generated multi-channel 3D face images. The input to the network are $182 \times 182 \times 9$ images where the nine channels correspond to the R, G, B colors, the three dimensions of a normal vector, the Gaussian curvature, the conformal factor, and the depth information respectively. Figure~\ref{fig:architecture} (b) shows an example of the 9-channel image. From the training set, we randomly select $90\%$ scans of each identity for training and use the remaining scans for validation.

\section{Experiments}\label{sec:experiments}

\subsection{3D Face Databases}
We report face recognition performances on the two standard public 3D databases: Bosphorus~\cite{Bosphorus} and TexasFRD~\cite{TexasFRD}. We randomly choose $15\%$ from the combined two databases for testing. The remaining is for training and validation. 

\begin{itemize}
	\item The Bosphorus database contains $4,666$ 3D facial scans with rich expression variations, poses, and occlusions from $105$ subjects generated by a stereo scanner.
	
	\item The TexasFRD database contains $1,151$ 3D facial scans with rich expression variations from $118$ subjects generated by a stereo scanner.
\end{itemize}

\subsection{Performance}

\begin{figure}[h!]
	\centering
	\begin{tabular}{cc}
		\includegraphics[height=6.2cm]{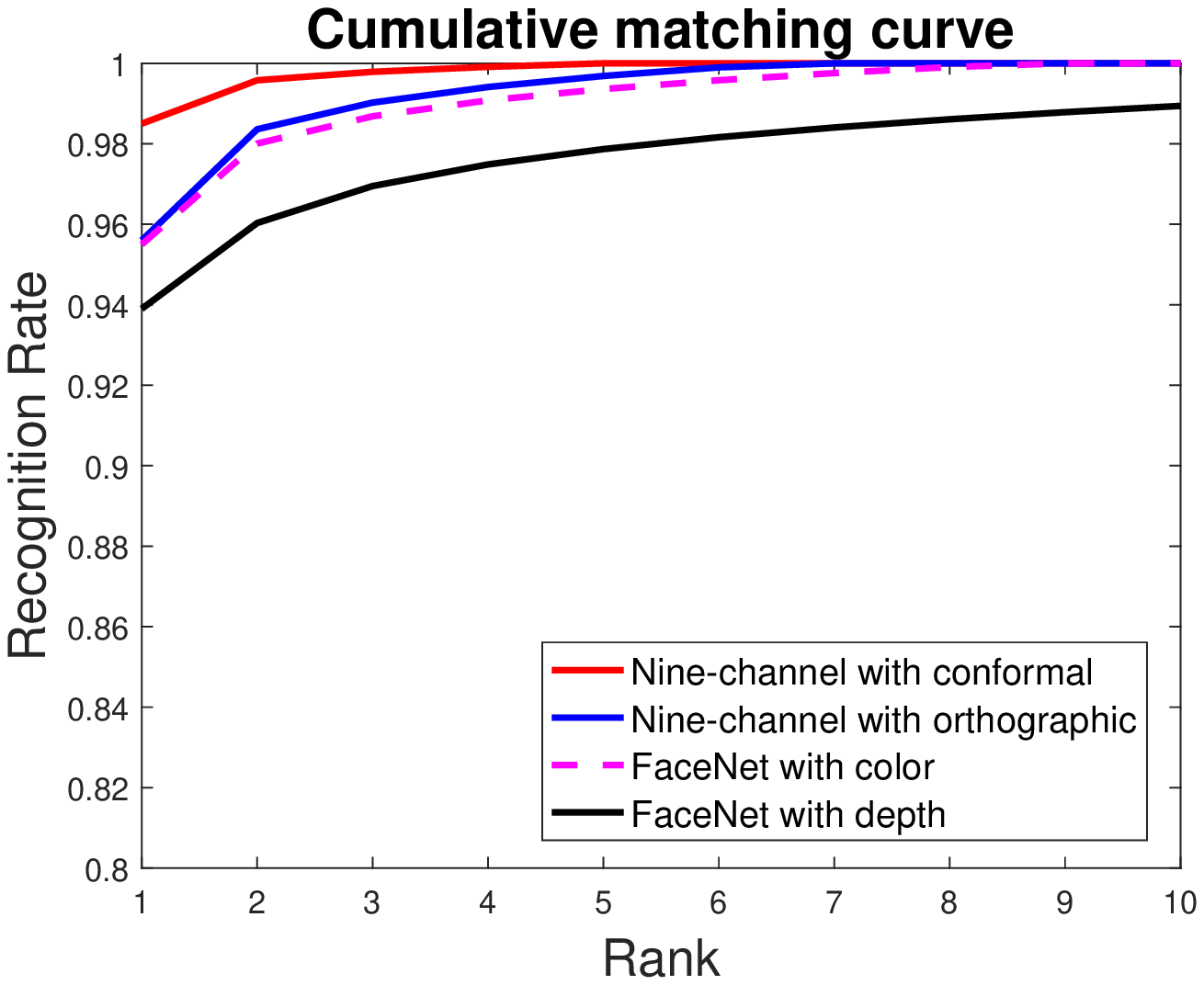} &
		\includegraphics[height=6.2cm]{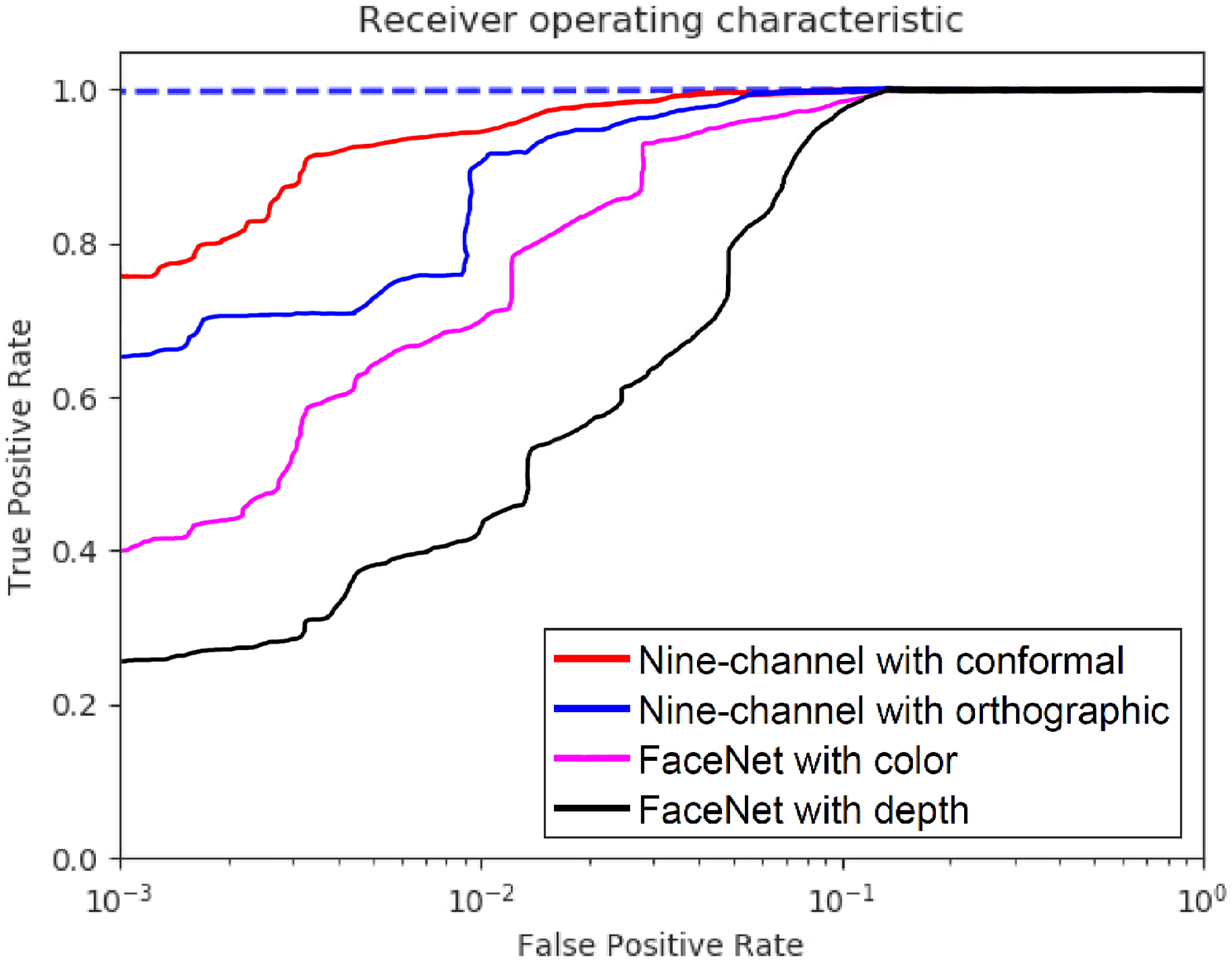} \\
		(a) CMC curve &	(b) ROC curve \\
	\end{tabular}
	\caption{ Comparison of face recognition: FaceNet with  the face color image as input; FaceNet with  the depth image as input; multi-Channel deep 3D face network with  the 9-channel image ``flattened'' to plane based on orthographic	projection as input; multi-Channel deep 3D face network with  the 9-channel image``flattened''to plane based on conformal mapping as input. }
	\label{fig:curve}
\end{figure}

We randomly select one scan of each identity from the testing set to place in the gallery while the remaining scans are used as probes. Face recognition is performed by computing the embedded L2 distance of a probe with all identities in the gallery. The identity with the smallest distance is assigned to the probe.

We compare the results with the state-of-the-art 2D face recognition network: FaceNet~\cite{FaceNet2015}. We feed FaceNet with the face color image and depth image, respectively. We also compare the results with a multi-Channel deep 3D face network where the nine-channel information is ``flattened'' to plane based on orthographic
projection. To perform the orthographic projection, we align all the facial 3D models together using a
classical rigid-ICP~\cite{Castellani_3dshape} between each 3D scan and a reference
facial model as~\cite{DBLP:journals/corr/KimHCM17}. 

We report the comparison results of recognition in the form of Cumulative Matching Curve (CMC)  as shown in Figure~\ref{fig:curve} (a)  and Receiver Operating Characteristic curve (ROC)  as shown in Figure~\ref{fig:curve} (b). Specifically, CMC provides face recognition precision for each rank. For each probe from the testing set, we sort the distances against the whole gallery and obtain the rank of the match. Face recognition performance is then stated as the fraction of probes whose gallery match is at rank $r$ or lower. It gives an estimation of the rate at which probe images will be successfully recognized at rank $r$ or better~\cite{Grother:2003}. ROC is computed where a varying threshold is applied to the L2 distance of the probe and each identity in the gallery.  It shows the trade-off between the true positive (true recognition)  and false positive (false recognition) rates as a parametric function of the prior distance threshold, where the true positive rate is the fraction of probes whose gallery match has distance smaller than or equal to the threshold value~\cite{Grother:2003}.

Table~\ref{tab:results} details the Rank-1 identification results of the multi-Channel deep 3D face network and compares with the state-of-the-art conventional and deep CNN based methods. Note that for those cited comparison methods we report the results from the original papers.

\begin{table*}
	\caption{Comparison of the Rank-1 identification results \label{tab:results}}\vspace{-6mm}
	\begin{center}
		%\ra{1.3}
		{    %\fontsize{9pt}{9pt}\selectfont
			\begin{tabular}[t]{ @{}  c | c | c | c |c  @{}} \toprule
				Method  & Model & Input & Bosphorus  & TexasFRD \\ \hline	
				\multirow{2}{*}{conventional} & MMH~\cite{MMH}  & 3D mesh + RGB image & 96.4\%    & 98.0\% \\
				& K3DM~\cite{K3DM}  & 3D mesh & 98.6\%    & 98.1\% \\ \hline
				\multirow{2}{*}{Deep CNN} & FaceNet  & RGB image &  95.5\%   &  95.5\% \\
				& FaceNet  & Depth  &  93.9\%   & 93.9\% \\ \hline
				\multirow{5}{*}{Deep CNN}  & DCNN~\cite{DBLP:journals/corr/KimHCM17}  & Depth &  99.2   & - \\ 
				& FR3DNet~\cite{DBLP:journals/corr/abs-1711-05942}  & Depth + Normal &  96.18\%   & 100\% \\
				& $FR3DNet_{FT}$~\cite{DBLP:journals/corr/abs-1711-05942}  & Depth + Normal &  100\%   & 100\% \\
				& Our Nework  & Nine channels (orthographic) &  95.6\%   & 95.6\% \\  
				& Our Network  & Nine channels (conformal) &  98.6\%   & 98.6\% \\ \hline
				\bottomrule
			\end{tabular}
		}
	\end{center}
\end{table*}

\subsection{Discussions and Limitation}

It is clear that the multi-Channel deep 3D face network performs much better when a 9-channel image is ``flattened'' to plane based on conformal mapping compared with orthographic projection.

The limitation is the dearth of labeled 3D face data for training the multi-Channel deep 3D face network. The authors in ~\cite{DBLP:journals/corr/KimHCM17} synthesize new 3D face data using multi-linear 3D morphable models such that their network is trained with depth images of $127K$ 3D scans of $700$ identities. The authors in~\cite{DBLP:journals/corr/abs-1711-05942} apply non-linear
heterogeneous variations in 3D shape, facial expressions,
pose and occlusions to generate a training dataset of $3.1M$
3D scans of $100K$ identities. 

It is obvious that without enough training data, the performance of the multi-Channel deep 3D face network is still below the two CNNs~\cite{DBLP:journals/corr/KimHCM17,DBLP:journals/corr/abs-1711-05942} , although the input to the network contains far more accurate geometric information produced by conformal mapping. 

\section{Conclusion and Future Works}\label{sec:conclusion}

We have designed a multi-channel deep 3D face network for face recognition.  We first compute the geometric information of a scanned 3D face. By employing a novel mathematical tool, discrete surface Ricci flow, we conformally ``flatten'' the geometric information and the face color from 3D  to 2D plane to leverage the state-of-the-art deep CNN architecture. We modify the input layer of the network to take images with nine channels such that more information can be explicitly fed to the network. Although the amount of the multi-channel face images is limited, the face recognition accuracy of the multi-channel deep 3D face network has achieved $98.6\%$. The experimental results clearly show that the network performs much better when a 9-channel image is ``flattened'' to plane based on a conformal map compared with orthographic projection.  As our future works, we will train the network on larger 3D database sets with automatically generated 3D face data.

\bibliographystyle{unsrtnat}
\bibliography{refs} 

\begin{thebibliography}{29}
\providecommand{\natexlab}[1]{#1}
\providecommand{\url}[1]{\texttt{#1}}
\expandafter\ifx\csname urlstyle\endcsname\relax
  \providecommand{\doi}[1]{doi: #1}\else
  \providecommand{\doi}{doi: \begingroup \urlstyle{rm}\Url}\fi

\bibitem[Parkhi et~al.(2015)Parkhi, Vedaldi, and Zisserman]{Parkhi2015}
O.~M. Parkhi, A.~Vedaldi, and A.~Zisserman.
\newblock Deep face recognition.
\newblock In \emph{In British Machine Vision Conference}, page~6, 2015.

\bibitem[Taigman et~al.(2014)Taigman, Yang, Ranzato, and Wolf]{DeepFace2014}
Yaniv Taigman, Ming Yang, Marc'Aurelio Ranzato, and Lior Wolf.
\newblock Deepface: Closing the gap to human-level performance in face
  verification.
\newblock In \emph{Proceedings of the 2014 IEEE Conference on Computer Vision
  and Pattern Recognition}, CVPR '14, pages 1701--1708, 2014.

\bibitem[Schroff et~al.(2015)Schroff, Kalenichenko, and Philbin]{FaceNet2015}
Florian Schroff, Dmitry Kalenichenko, and James Philbin.
\newblock Facenet: {A} unified embedding for face recognition and clustering.
\newblock In \emph{Proceedings of the 2015 IEEE Conference on Computer Vision
  and Pattern Recognition}, CVPR '15, page 815–823, 2015.

\bibitem[Ahonen et~al.(2004)Ahonen, Hadid, and
  Pietikäinen]{Ahonen_facerecognition2004}
Timo Ahonen, Abdenour Hadid, and Matti Pietikäinen.
\newblock Face recognition with local binary patterns.
\newblock In \emph{In European conference on computer vision}, pages 469--481,
  2004.

\bibitem[Simonyan et~al.(2013)Simonyan, Parkhi, Vedaldi, and
  Zisserman]{Simonyan13}
K.~Simonyan, O.~M. Parkhi, A.~Vedaldi, and A.~Zisserman.
\newblock {F}isher {V}ector {F}aces in the {W}ild.
\newblock In \emph{British Machine Vision Conference}, page~4, 2013.

\bibitem[Kim et~al.(2017)Kim, Hernandez, Choi, and
  Medioni]{DBLP:journals/corr/KimHCM17}
Donghyun Kim, Matthias Hernandez, Jongmoo Choi, and G{\'{e}}rard~G. Medioni.
\newblock Deep 3d face identification.
\newblock \emph{CoRR}, abs/1703.10714, 2017.
\newblock URL \url{http://arxiv.org/abs/1703.10714}.

\bibitem[Gilani and Mian(2018)]{DBLP:journals/corr/abs-1711-05942}
Syed~Zulqarnain Gilani and Ajmal Mian.
\newblock Learning from millions of 3d scans for large-scale 3d face
  recognition.
\newblock In \emph{IEEE Conference of Computer Vision and Pattern Recognition
  (CVPR)}, 2018.

\bibitem[Savran et~al.(2008)Savran, Aly\"{u}z, Dibeklio\u{g}lu, \c{C}eliktutan,
  G\"{o}kberk, Sankur, and Akarun]{Bosphorus}
Arman Savran, Ne\c{s}e Aly\"{u}z, Hamdi Dibeklio\u{g}lu, Oya \c{C}eliktutan,
  Berk G\"{o}kberk, B\"{u}lent Sankur, and Lale Akarun.
\newblock Biometrics and identity management.
\newblock chapter Bosphorus Database for 3D Face Analysis, pages 47--56. 2008.

\bibitem[Yin et~al.(2006)Yin, Wei, Sun, Wang, and Rosato]{BU3DFE}
Lijun Yin, Xiaozhou Wei, Yi~Sun, Jun Wang, and Matthew~J. Rosato.
\newblock A 3d facial expression database for facial behavior research.
\newblock In \emph{Proceedings of the 7th International Conference on Automatic
  Face and Gesture Recognition}, FGR '06, pages 211--216, 2006.

\bibitem[Vijayan et~al.(2011)Vijayan, Bowyer, Flynn, Huang, Chen, Hansen,
  Ocegueda, Shah, and Kakadiaris]{3D-TEC}
V.~Vijayan, K.~W. Bowyer, P.~J. Flynn, D.~Huang, L.~Chen, M.~Hansen,
  O.~Ocegueda, S.~K. Shah, and I.~A. Kakadiaris.
\newblock Twins 3d face recognition challenge.
\newblock In \emph{International Joint Conference on Biometrics (IJCB)}, pages
  1--7, 2011.

\bibitem[Hamilton(1982)]{ric82}
Richard~S. Hamilton.
\newblock Three manifolds with positive {R}icci curvature.
\newblock \emph{Journal of Differential Geometry.}, 17:\penalty0 255--306,
  1982.

\bibitem[Chow and Luo(2003)]{chow_luo_03}
Bennett Chow and Feng Luo.
\newblock {Combinatorial {R}icci Flows on Surfaces}.
\newblock \emph{Journal Differential Geometry}, 63\penalty0 (1):\penalty0
  97--129, 2003.

\bibitem[Jin et~al.(2008)Jin, Kim, Luo, and Gu]{TVCG_08_Ricci}
Miao Jin, Junho Kim, Feng Luo, and Xianfeng Gu.
\newblock Discrete surface ricci flow.
\newblock \emph{IEEE Transactions on Visualization and Computer Graphics},
  14\penalty0 (5):\penalty0 1030--1043, 2008.

\bibitem[Thurston(1976)]{Thurston76}
William~P. Thurston.
\newblock \emph{Geometry and Topology of Three-Manifolds}.
\newblock Princeton lecture notes, 1976.

\bibitem[Abbena et~al.(2006)Abbena, Salamon, and Gray]{Gray:2006}
Elsa Abbena, Simon Salamon, and Alfred Gray.
\newblock \emph{Modern Differential Geometry of Curves and Surfaces with
  Mathematica}.
\newblock CRC Press, Inc., 3rd edition, 2006.
\newblock ISBN 1584884487.

\bibitem[Meyer et~al.(2002)Meyer, Desbrun, Schröder, and
  Barr]{Meyer02discretedifferential-geometry}
Mark Meyer, Mathieu Desbrun, Peter Schröder, and Alan~H. Barr.
\newblock Discrete differential-geometry operators for triangulated
  2-manifolds, 2002.

\bibitem[Cohen-Steiner and Morvan(2003)]{Cohen-Steiner:2003:RDT:777792.777839}
David Cohen-Steiner and Jean-Marie Morvan.
\newblock Restricted delaunay triangulations and normal cycle.
\newblock In \emph{Proceedings of the Nineteenth Annual Symposium on
  Computational Geometry}, SCG '03, pages 312--321, 2003.
\newblock ISBN 1-58113-663-3.

\bibitem[Rusinkiewicz(2004)]{Rusinkiewicz:2004:ECD:1018408.1018660}
Szymon Rusinkiewicz.
\newblock Estimating curvatures and their derivatives on triangle meshes.
\newblock In \emph{Proceedings of the 3D Data Processing, Visualization, and
  Transmission, 2Nd International Symposium}, 3DPVT '04, pages 486--493, 2004.
\newblock ISBN 0-7695-2223-8.

\bibitem[Gatzke and Grimm(2006)]{Gatzke06estimatingcurvature}
T.~D. Gatzke and C.~M. Grimm.
\newblock Estimating curvature on triangular meshes.
\newblock \emph{International Journal of Shape Modeling}, 2006.

\bibitem[Zhang et~al.(2016)Zhang, Zhang, Li, and
  Qiao]{DBLP:journals/corr/ZhangZL016}
Kaipeng Zhang, Zhanpeng Zhang, Zhifeng Li, and Yu~Qiao.
\newblock Joint face detection and alignment using multi-task cascaded
  convolutional networks.
\newblock \emph{CoRR}, abs/1604.02878, 2016.
\newblock URL \url{http://arxiv.org/abs/1604.02878}.

\bibitem[He et~al.(2016)He, Zhang, Ren, and Sun]{DBLP:conf/cvpr/HeZRS16}
Kaiming He, Xiangyu Zhang, Shaoqing Ren, and Jian Sun.
\newblock Deep residual learning for image recognition.
\newblock In \emph{2016 {IEEE} Conference on Computer Vision and Pattern
  Recognition, {CVPR} 2016, Las Vegas, NV, USA, June 27-30, 2016}, pages
  770--778, 2016.

\bibitem[Russakovsky et~al.(2015)Russakovsky, Deng, Su, Krause, Satheesh, Ma,
  Huang, Karpathy, Khosla, Bernstein, Berg, and
  Fei-Fei]{Russakovsky:2015:ILS:2846547.2846559}
Olga Russakovsky, Jia Deng, Hao Su, Jonathan Krause, Sanjeev Satheesh, Sean Ma,
  Zhiheng Huang, Andrej Karpathy, Aditya Khosla, Michael Bernstein,
  Alexander~C. Berg, and Li~Fei-Fei.
\newblock Imagenet large scale visual recognition challenge.
\newblock \emph{Int. J. Comput. Vision}, 115\penalty0 (3):\penalty0 211--252,
  2015.

\bibitem[Ioffe and Szegedy(2015)]{Ioffe:2015:BNA:3045118.3045167}
Sergey Ioffe and Christian Szegedy.
\newblock Batch normalization: Accelerating deep network training by reducing
  internal covariate shift.
\newblock In \emph{Proceedings of the 32Nd International Conference on
  International Conference on Machine Learning - Volume 37}, pages 448--456,
  2015.

\bibitem[He et~al.(2015)He, Zhang, Ren, and Sun]{He:2015:DDR:2919332.2919814}
Kaiming He, Xiangyu Zhang, Shaoqing Ren, and Jian Sun.
\newblock Delving deep into rectifiers: Surpassing human-level performance on
  imagenet classification.
\newblock In \emph{Proceedings of the 2015 IEEE International Conference on
  Computer Vision (ICCV)}, ICCV '15, pages 1026--1034, 2015.

\bibitem[Gupta et~al.(2010)Gupta, Markey, and Bovik]{TexasFRD}
Shalini Gupta, Mia~K. Markey, and Alan~C. Bovik.
\newblock Anthropometric 3d face recognition.
\newblock \emph{Int. J. Comput. Vision}, 90\penalty0 (3):\penalty0 331--349,
  December 2010.
\newblock ISSN 0920-5691.

\bibitem[Castellani and Bartoli(2012)]{Castellani_3dshape}
Umberto Castellani and Adrien Bartoli.
\newblock 3d shape registration.
\newblock \emph{3D Imaging, Analysis and Applications, Springer}, page
  221–264, 2012.

\bibitem[Grother et~al.(2003)Grother, Micheals, and Phillips]{Grother:2003}
Patrick Grother, Ross~J. Micheals, and P.~Jonathon Phillips.
\newblock Face recognition vendor test 2002 performance metrics.
\newblock \emph{Audio-and Video-Based Biometric Person Authentication. Springer
  Berlin Heidelberg}, 2003.

\bibitem[Mian et~al.(2007)Mian, Bennamoun, and Owens]{MMH}
A.~Mian, M.~Bennamoun, and R.~Owens.
\newblock An efficient multimodal 2d-3d hybrid approach to automatic face
  recognition.
\newblock \emph{IEEE Transactions on Pattern Analysis and Machine
  Intelligence}, 29\penalty0 (11):\penalty0 1927–1943, 2007.

\bibitem[Gilani et~al.(2018)Gilani, Mian, Shafait, and Reid]{K3DM}
S.~Z. Gilani, A.~Mian, F.~Shafait, and I.~Reid.
\newblock Dense 3d face correspondence.
\newblock \emph{IEEE Transactions on Pattern Analysis and Machine
  Intelligence}, 40\penalty0 (7):\penalty0 1584 -- 1598, December 2018.

\end{thebibliography}

\end{document}